\documentclass[runningheads]{llncs}

\usepackage{amssymb}
\usepackage{graphicx}
\usepackage{times}
\usepackage[tbtags]{amsmath}
\usepackage{mathtools}
\usepackage{enumerate}
\usepackage{tikz}
\usepackage{url}
\usepackage{graphicx}
\usepackage{multirow}
\usepackage{nth}
\usepackage{cite}
\usepackage{longtable}
\usepackage{caption}
\usepackage{url}
\usetikzlibrary{trees,shapes,arrows.meta}
\usepackage{enumitem}
\setitemize{topsep=0.1em,parsep=0.1em,partopsep=0.1em}

\usepackage{array}
\newcolumntype{L}[1]{>{\raggedright\let\newline\\\arraybackslash\hspace{0pt}}m{#1}}
\newcolumntype{C}[1]{>{\centering\let\newline\\\arraybackslash\hspace{0pt}}m{#1}}
\newcolumntype{R}[1]{>{\raggedleft\let\newline\\\arraybackslash\hspace{0pt}}m{#1}}

\usepackage{algorithm}
\usepackage{algpseudocode}

\newcommand{\tp}{\textit{TP}}
\newcommand{\fp}{\textit{FP}}
\newcommand{\tn}{\textit{TN}}
\newcommand{\fn}{\textit{FN}}

\newcommand{\heur}{\ensuremath{h}}
\newcommand{\labelfont}[1]{\emph{#1}}
\newcommand{\labelfontinv}[1]{\ensuremath{\overline{\text{\labelfont{#1}}}}}

\algloopdefx{NoEndIf}[1]{\textbf{if} #1 \textbf{then}}

\begin{document}

\title{Efficient Discovery of Expressive Multi-label Rules using Relaxed Pruning}

\author{
\textit{\textbf{Preprint version}. To appear in Proceedings of the 22nd International Conference on Discovery Science, 2019}\\ \ \\ 
Yannik Klein  \and Michael Rapp \and Eneldo Loza Menc\'ia}
\authorrunning{Y. Klein  \and M. Rapp \and E. Loza Menc\'ia
}
\institute{Knowledge Engineering Group, TU Darmstadt, Germany\\ \email{yannik.klein@hotmail.com}, 
\email{\{mrapp,eneldo\}@ke.tu-darmstadt.de}}

\maketitle

\begin{abstract}
Being able to model correlations between labels is considered crucial in multi-label classification. Rule-based models enable to expose such dependencies, e.g., implications, subsumptions, or exclusions, in an interpretable and human-comprehensible manner. Albeit the number of possible label combinations increases exponentially with the number of available labels, it has been shown that rules with multiple labels in their heads, which are a natural form to model local label dependencies, can be induced efficiently by exploiting certain properties of rule evaluation measures and pruning the label search space accordingly. However, experiments have revealed that multi-label heads are unlikely to be learned by existing methods due to their restrictiveness. To overcome this limitation, we propose a plug-in approach that relaxes the search space pruning used by existing methods in order to introduce a bias towards larger multi-label heads resulting in more expressive rules. We further demonstrate the effectiveness of our approach empirically and show that it does not come with drawbacks in terms of training time or predictive performance.

\end{abstract}

\section{Introduction}
\label{section_introduction}

As many real world problems require to assign a set of labels, rather than a single class, to instances, multi-label classification (MLC) has become an established topic in the recent machine learning literature. For example, text documents are often related to multiple subjects and media, such as images or music, can usually be associated with several tags at the same time (see \cite{tsoumakas10MLoverview} for an overview).

Rule-based methods are a well-researched approach to solve classification problems. Due to their interpretability, the use of rule learning algorithms in MLC has recently been proposed as an alternative to complex statistical methods such as support vector machines or artificial neural networks (see e.g. \cite{MLRLbook, lakkaraju2016}). Rules provide a natural and simple representation of a learned model and can easily be understood, analyzed, and modified by human domain experts. Especially in safety-critical domains, such as medicine, power systems, or financial markets, the interpretability of machine learning models is an important requirement to be able to prevent malfunctions and unexpected behavior. 

Rules do not only reveal patterns and regularities hidden in the data, but are also able to make \emph{global} or \emph{local} correlations between labels explicit \cite{menc15}. Exploiting such correlations is considered crucial in MLC and it is commonly accepted that approaches that are able to take label dependencies into account can be expected to achieve better predictive results \cite{tsoumakas10MLoverview, demb12, menc15}. Existing multi-label rule learning approaches are able to exploit label correlations by inducing \emph{label-dependent} rules, i.e., rules that may contain label conditions in their bodies \cite{menc15}. In addition, rules with \emph{multi-label heads} provide the ability to model local dependencies between labels by including multiple label assignments in their heads \cite{rapp2018}. This enables to model co-occurrences --- a frequent pattern in multi-label data ---, as well as other types of interdependencies between labels, in a natural and compact form.

\subsubsection{Motivation and Goals.}
\label{section_motivation}

The induction of multi-label heads is particularly challenging as the number of label combinations that can potentially be included in a head increases exponentially with the number of available labels. To mitigate the computational complexity that comes with a search for multi-label heads, certain properties of commonly used multi-label evaluation measures --- namely \emph{anti-monotonicity} and \emph{decomposability} --- have successfully been exploited for pruning the search space. Although this enables to efficiently induce multi-label heads in theory, experiments have revealed that such patterns are unlikely to be learned in practice \cite{rapp2018}. This is due to the restrictiveness of existing methods that assess the quality of potential heads solely in terms of the used evaluation function. These functions tend to prefer single-label predictions to rules with multi-label heads, because the quality of the individual label assignments contained in such a head usually varies. For example, if two rules with the same body but different  single-label heads reach a heuristic value of 0.89 and 0.88, respectively, predicting both labels usually results in a performance decline compared to the value 0.89 --- typically a value between 0.89 and 0.88. However, opting for the multi-label head would arguably be a good choice: First, the resulting rule would have greater coverage. Second, it evaluates to a heuristic value only slightly worse than that of the best single-label rule. 

In this work, we present a relaxed pruning strategy to overcome the bias towards single-label predictions. We further argue that strict upper bounds in terms of computational complexity can still be guaranteed when relaxing the search for multi-label heads. As our empirical studies reveal, the training process even tends to terminate earlier due to the increased coverage of the induced rules. The experiments also show that the use of relaxed pruning results in more compact models that reach predictive results comparable to those of existing approaches. Moreover, we discuss whether our approach discovers more label dependencies, which is a major goal of our method.

\subsubsection{Structure of this Work.} \label{section_structure}

We start with an introduction to the problem domain and a recapitulation of previous work in Section~\ref{section_preliminaries}. As the main contribution of this work, in Sections~\ref{section_relaxed_pruning} and \ref{section:relaxed_search_space_pruning}, we present a plug-in approach that relaxes the search space pruning used by existing methods in order to introduce a bias towards the induction of larger multi-label heads. To illustrate the effects of our extension, we present an empirical analysis focusing on the predictive performance, model characteristics and runtime efficiency of the proposed method in Section~\ref{section_evaluation}. Finally, we provide an overview of related work in Section~\ref{section_related_work} before we conclude by summarizing our results in Section~\ref{section_conclusions}.

\section{Preliminaries} 
\label{section_preliminaries}

In contrast to binary or multi-class classification, in MLC an instance can be associated with several labels $\lambda_i$ out of a predefined label space $\mathbb{L} = \{ \lambda_1, \dots, \lambda_n \}$. The task is to learn a classifier function $g \left( . \right)$ that maps an instance $\boldsymbol{x}$ to a predicted label vector $\boldsymbol{\hat{y}} = \left( \hat{y}_1, \dots , \hat{y}_n \right) = \{ 0, 1 \}^n$, where each prediction $\hat{y}_i$ specifies the presence (1) or absence (0) of the corresponding label $\lambda_i$. An instance $\boldsymbol{x}_j$ consists of attribute-value pairs given as a vector $\boldsymbol{x}_j = \left( v_1, \dots, v_l \right) \in \mathbb{D} = A_1 \times \dots \times A_l$, where $A_i$ is a numeric or nominal attribute. We handle MLC as a supervised learning problem in which the classifier function $g \left( . \right)$ is induced from labeled training data $T = \{ \left( \boldsymbol{x}_1, \boldsymbol{y}_1 \right), \dots, \left( \boldsymbol{x}_m, \boldsymbol{y}_m \right) \}$, containing tuples of training instances $\boldsymbol{x}_j$ and true label vectors $\boldsymbol{y}_j$.

\subsection{Multi-label Classification Rules}
\label{section_multi_label_rules}

We are concerned with the induction of conjunctive, propositional rules $\boldsymbol{r}: \hat{Y} \leftarrow B$. On the one hand, the body $B$ of such a rule contains an arbitrary number of conditions that compare an attribute-value $v_i$ of an instance to a constant by using equality (nominal attributes) or inequality (numerical attributes) operators. If an instance satisfies all conditions in the body of a rule $\boldsymbol{r}$, it is said to be \emph{covered} by $\boldsymbol{r}$. 
On the other hand, the head $\hat{Y}$ consists of one (\emph{single-label head}) or several (\emph{multi-label head}) label assignments $\hat{y}_i = \{0, 1\}$ that specify whether the label $\lambda_i$ should be predicted as present (1) or absent (0) for the covered instances. Multi-label heads enable to model local dependencies, such as co-occurrences or exclusions, that hold for the instance subspace covered by the rule's body.

In general, the head $\hat{Y}$ of a rule may have different semantics in a multi-label setting. We consider the predictions provided by an individual rule to be \emph{partial}, because we believe that this particular strategy has several conceptual and practical advantages. When using partial predictions, each rule only predicts the presence or absence of a subset of the available labels and leaves the prediction of the remaining ones to other rules.

\subsection{Bipartition Evaluation Functions}
\label{section_bipartition_functions}

To evaluate the quality of multi-label predictions, we use bipartition evaluation functions $\delta: \mathbb{N}^{2 \times 2} \rightarrow \mathbb{R}$ that are based on comparing the difference between true label vectors (\emph{ground truth}) and predicted labels (cf.~\cite{tsoumakas10MLoverview}). Such a function maps a two-dimensional (label) confusion matrix $C$ to a heuristic value $h \in \left[ 0, 1 \right]$. A confusion matrix consists of the number of \emph{true positive} ($\tp$), \emph{false positive} ($\fp$), \emph{true negative} ($\tn$), and \emph{false negative} ($\fn$) labels predicted by a rule or classifier.

Let the variables $y_i^j$ and $\hat{y}_i^j$ denote the absence (0) or presence (1) of the label $\lambda_i$ of instance $\boldsymbol{x}_j$ according to the ground truth or a prediction, respectively. Given these variables, we calculate the atomic confusion matrix $C_i^j$ for the respective label $\lambda_i$ and instance $\boldsymbol{x}_j$ as
\begin{equation}
  \label{equation_confusion_matrix}
  C_i^j = \left(\begin{matrix}
    \tp_i^j & \fp_i^j \\
    \fn_i^j & \tn_i^j
  \end{matrix}\right)
= \left( \begin{matrix}
  y_{i}^j  \hat{y}_{i}^j &&
(1-y_{i}^j)  \hat{y}_{i}^j \\
 (1-y_{i}^j)  (1-\hat{y}_{i}^j) &&
 y_{i}^j  (1-\hat{y}_{i}^j)
\end{matrix}\right)
\end{equation}

Note that, in accordance with \cite{rapp2018}, we assess $\tp$, $\fp$, $\tn$, and $\fn$ differently to evaluate candidate rules during training. To ensure that absent and present labels have the same impact on the performance of a rule, we always count correctly predicted labels as $\tp$ and incorrect predictions as $\fp$. Labels for which no prediction is made are counted as $\tn$ if they are absent, or as $\fn$ if they are present.

When evaluating multi-label predictions which have been made for $m$ instances and $n$ labels it is necessary to aggregate the resulting $m \cdot n$ atomic confusion matrices. We restrict ourselves to \emph{micro-} and \emph{(label-based) macro-averaging}, which are defined as
\begin{equation}
  \label{equation_micro_macro_averaging}
  \delta \left( C \right) = \delta \left( \sum\nolimits_i \sum\nolimits_j C_i^j \right)\quad
  \text{and}\quad
  \delta \left( C \right) = \text{avg}_i \left( \delta \left( \sum\nolimits_j C_i^j \right) \right)
\end{equation}
where the $\sum$ operator denotes the cell-wise addition of atomic confusion matrices $C_i^j$, corresponding to label $\lambda_i$ and instance $\boldsymbol{x}_j$, and $\text{avg}_i$ calculates the mean of the heuristic values obtained for each label $\lambda_i$.

\subsection{Multi-label Rule Learning Heuristics}
\label{section_heuristics}

In the following, we present the bipartition evaluation functions --- also referred to as \emph{heuristics} --- that are used in this work to assess the quality of candidate rules in terms of a heuristic value $h$. According to these heuristics, rules with a greater heuristic value are preferred to those with smaller values.

Among the heuristics we use in this work is \emph{Hamming accuracy} (HA). It measures the percentage of correctly predicted labels among all labels and can be computed using micro and macro-averaging with the same final result.
\begin{equation}
  \label{equation_hamming_accuracy}
  \delta_\textit{hamm} \left( C \right) \coloneqq \frac{\tp + \tn}{\tp + \fp + \tn + \fn}
\end{equation}

Moreover, we use the \emph{F-measure} (FM) to evaluate candidate rules. It calculates as the (weighted) harmonic mean of \emph{precision} and \emph{recall}. If $\beta < 1$, precision has a greater impact. If $\beta > 1$, the F-measure becomes more recall-oriented.
\begin{equation}
    \label{equation_f_measure}
    \delta_{F} \left( C \right) \coloneqq \frac{\left( 1 + \beta^2 \right) \cdot \tp}{\left( 1 + \beta^2 \right) \cdot \tp + \beta^2 \cdot \fn + \fp} \text{ with } \beta \in \left[ 0, +\infty \right]
\end{equation}

\subsection{Pruning the Search for Multi-label Heads}
\label{section_search_space_pruning}

We rely on the multi-label rule learning algorithm proposed by Rapp~et~al.~\cite{rapp2018} to learn rule-based models. It uses a \emph{separate-and-conquer} strategy, where new rules are induced iteratively. Whenever a new rule is learned, the covered instances are removed from the training data set if enough of their labels are predicted by already induced rules (``separate'' step). Afterwards, the next rule is induced from the remaining instances and labels (``conquer'' step). The training process continues until only few training instances are left. To classify an instance, the rules contained in the model are applied in the order of their induction. If a rule covers the given instance, the labels in its head are applied unless they were already predicted by a previous rule.

To learn new rules, the algorithm performs a top-down greedy search, starting with the most general rule. By adding additional conditions to the rule's body it is successively specialized, resulting in fewer instances being covered. For each candidate rule, a corresponding single- or multi-label head, that models the labels of the covered instances as accurately as possible, must be found.

To find a suitable (multi-label) head for a given body, potential label combinations are evaluated with respect to a certain heuristic $\delta$ using a breadth-first search. Instead of performing an exhaustive search through the label space, which is unfeasible in practice due to its exponential complexity, the search is pruned by leaving out unpromising label combinations as illustrated in Figure~\ref{figure_example}. To prune the search for multi-label heads, while still being able to find the best solution, Rapp~et~al.~\cite{rapp2018} exploit certain properties of multi-label evaluation measures --- namely \emph{anti-monotonicity} and \emph{decomposability}. In this work, we focus on the latter for two reasons: First, decomposability is a stronger criterion compared to anti-monotonicity. It enables to prune the search space more extensively and comes with linear costs, i.e., the best multi-label head can be inferred from considering each label separately. Second, most common multi-label evaluation measures have been proved to be decomposable, including the ones used in this work (cf.~Section~\ref{section_heuristics}). The definition of decomposability is given below.

\begin{definition}[Decomposability, cf. \cite{rapp2018}]
\label{definition_decomposability}
A multi-label evaluation function $\delta$ is \emph{decomposable} if the following conditions are met:
\begin{enumerate}[label=\emph{\roman*})]
  \item If the multi-label head rule $\hat{Y} \leftarrow B$ contains a label attribute $\hat{y}_{i} \in \hat{Y}$ for which the corresponding single-label head rule $\hat{y}_{i} \leftarrow B$ does not reach $h_{max}$, the multi-label head rule cannot reach that performance either (and vice versa).
\[
\exists i \left( \hat{y}_{i} \in \hat{Y} \wedge \heur(\hat{y}_{i} \leftarrow B) < h_{max} \right) \Longleftrightarrow \heur(\hat{Y} \leftarrow B) < h_{max}
\]
  \item If all single label head rules $\hat{y}_{i} \leftarrow B$ which correspond to the label attributes of the multi-label head $\hat{Y}$ reach $h_{max}$, the multi-label head rule $\hat{Y} \leftarrow B$ reaches that performance as well (and vice versa).
\[
\heur(\hat{y}_{i} \leftarrow B) = h_{max} \text{ , } \forall \hat{y}_{i} \left( \hat{y}_{i} \in \hat{Y} \right) \Longleftrightarrow \heur(\hat{Y} \leftarrow B) = h_{max}
\]
\end{enumerate}
\end{definition}

According to Definition~\ref{definition_decomposability}, we can safely prune the search space by restricting the evaluation to all possible single-label heads for a given body. To construct the best multi-label head, the highest heuristic value among all single-label heads is determined and those achieving the highest value are combined, while the others are discarded.

\tikzstyle{level 1}=[level distance=1.3cm, sibling distance=3.8cm]
\tikzstyle{level 2}=[level distance=1.7cm, sibling distance=2.2cm]
\tikzstyle{level 3}=[level distance=1.9cm, sibling distance=2.1cm]
\tikzstyle{level 4}=[level distance=1.9cm]
\tikzstyle{bag} = [text width=5.0em, text centered]
\tikzstyle{circled-bag} = [bag, circle, dashed, draw=black, inner sep=0pt]
\tikzstyle{arrow} = [-{Latex[scale=1.5]}]
\begin{figure}[t]
\centering
\resizebox{\textwidth}{!}{
\begin{tikzpicture}
\node[bag] {$\boldsymbol{\emptyset}$}
  child[black,growth parent anchor={west}] {
    node(a2)[bag,black] {$\boldsymbol{\{\hat{y}_{1}\}}$ \\ $h=\hat{h}=\frac{2}{3}$}
    child[red] {
      node[circled-bag,black] {$\boldsymbol{\{\hat{y}_{1},\hat{y}_{2}\}}$ \\ $h=\frac{2}{3}$ $\hat{h}=\frac{11}{15}$}
      child[red,growth parent anchor={center}] {
        node(a)[bag,black] {$\boldsymbol{\{\hat{y}_{1},\hat{y}_{2},\hat{y}_{3}\}}$ \\ $h=\frac{5}{9}$ $\hat{h}=\frac{23}{36}$}
          child[red] {
            node[bag,black] {$\boldsymbol{\{\hat{y}_{1},\hat{y}_{2},\hat{y}_{3},\hat{y}_{4}\}}$ \\ $h=\frac{5}{12}$ $\hat{h}=\frac{119}{240}$}
            edge from parent[arrow]
          }
        edge from parent[arrow]
      }
      child[red] {
        node(b)[bag,black] {$\boldsymbol{\{\hat{y}_{1},\hat{y}_{2},\hat{y}_{4}\}}$ \\ $h=\frac{4}{9}$}
        edge from parent[arrow]
      }
      edge from parent[arrow]
    }
    child[red,growth parent anchor={center}] {
      node(c)[bag,black] {$\boldsymbol{\{\hat{y}_{1},\hat{y}_{3}\}}$ \\ $h=\frac{1}{2}$}
      child[red] {
        node[bag,black] {$\boldsymbol{\{\hat{y}_{1},\hat{y}_{3},\hat{y}_{4}\}}$ \\ $h=\frac{1}{3}$}
        edge from parent[arrow]
      }
      edge from parent[arrow]
    }
    child[red] {
      node[bag,black] {$\boldsymbol{\{\hat{y}_{1},\hat{y}_{4}\}}$ \\ $h=\frac{1}{3}$}
      edge from parent[arrow]
    }
    edge from parent[arrow]
  }
  child[black] {
    node[bag,black] {$\boldsymbol{\{\hat{y}_{2}\}}$ \\ $h=\hat{h}=\frac{2}{3}$}
    child[red] {
      node[bag,black] {$\boldsymbol{\{\hat{y}_{2},\hat{y}_{3}\}}$ \\ $h=\frac{1}{2}$}
        child[red] {
          node[bag,black] {$\boldsymbol{\{\hat{y}_{2},\hat{y}_{3},\hat{y}_{4}\}}$ \\ $h=\frac{1}{3}$}
          edge from parent[arrow]
        }
      edge from parent[arrow]
    }
    child[red] {
      node(d)[bag,black] {$\boldsymbol{\{\hat{y}_{2},\hat{y}_{4}\}}$ \\ $h=\frac{1}{3}$}
      edge from parent[arrow]
    }
    edge from parent[arrow]
  }
  child[black,xshift=-1.0cm] {
    node[bag,black] {$\boldsymbol{\{\hat{y}_{3}\}}$ \\ $h=\hat{h}=\frac{1}{3}$}
    child[red] {
      node(e)[bag,black] {$\boldsymbol{\{\hat{y}_{3},\hat{y}_{4}\}}$ \\ $h=\frac{1}{6}$}
      edge from parent[arrow]
    }
    edge from parent[arrow]
  }
  child[black,xshift=-2.9cm,yshift=+0.06cm] {
    node(b2)[bag,black,yshift=+0.0cm] {$\boldsymbol{\{\hat{y}_{4}\}}$ \\ $h=\hat{h}=0$}
    edge from parent[arrow]
  };
  \draw[dashed] ([yshift=-0.1cm] a.south west) to ([xshift=0.6cm, yshift=-0.1cm] a.south) to[out=0,in=180] ([xshift=-2.6cm, yshift=-0.6cm] c.south) to ([xshift=-1.5cm, yshift=-0.6cm] c.south) to[out=0,in=180] ([yshift=-0.4cm, xshift=-0.50cm] a2.south west) to  ([xshift=1.5cm, yshift=-0.5cm] b2.south west);
  \draw[solid] ([xshift=-1.0cm, yshift=-0.2cm] a2.south west) to ([xshift=+1.5cm, yshift=-0.2cm] b2.south west |-  a2.south east);

  \node(table) at ([xshift=-0.4cm,yshift=-2.3cm] e.south east) {
    \begin{tabular}{c c|c c c c|}
    \cline{3-6}
    & & $\lambda_{1}$ & $\lambda_{2}$ & $\lambda_{3}$ & $\lambda_{4}$ \\
    \hline
    \multicolumn{1}{|c|}{\multirow{3}{*}{Not covered}} & $Y_{1}$ & 0 & 1 & 1 & 0 \\
    \multicolumn{1}{|c|}{} & $Y_{2}$ & 1 & 1 & 1 & 1 \\
    \multicolumn{1}{|c|}{} & $Y_{3}$ & 0 & 0 & 1 & 0 \\
    \hline
    \multicolumn{1}{|c|}{\multirow{3}{*}{Covered}} & $Y_{4}$ & 0 & 1 & 1 & 0 \\
    \multicolumn{1}{|c|}{} & $Y_{5}$ & 1 & 1 & 0 & 0 \\
    \multicolumn{1}{|c|}{} & $Y_{6}$ & 1 & 0 & 0 & 0 \\
    \hline
    \end{tabular}
  };
\end{tikzpicture}
}
\caption{Search for the best (relaxed) multi-label head given the labels $\lambda_1$, $\lambda_2$, $\lambda_3$, and $\lambda_4$. The instances corresponding to the label sets $Y_4$, $Y_5$, and $Y_6$ are assumed to be covered, whereas those of $Y_1$, $Y_2$, and $Y_3$ are not. The dashed line indicates label combinations that can be pruned with relaxed pruning, the solid line corresponds to standard decomposability (cf.~\cite{rapp2018},~Fig.~1).}
\label{figure_example}
\end{figure}
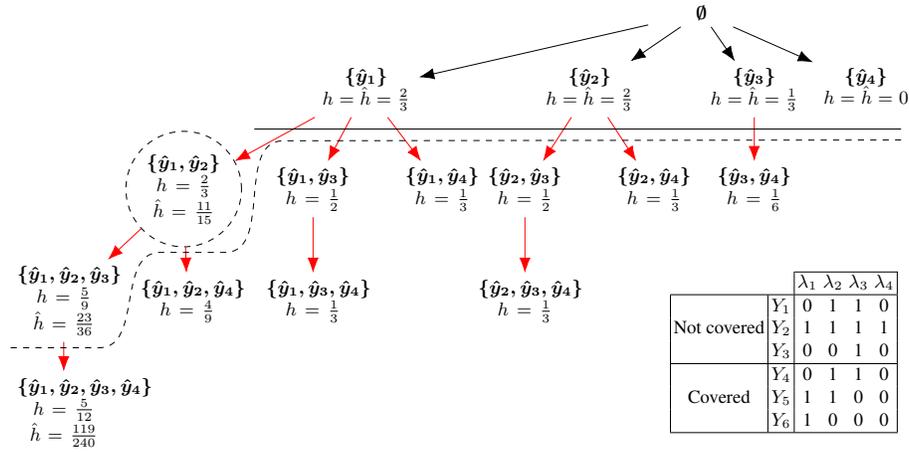

\section{Dynamic Weighting of Rules using Relaxation Lift Functions}
\label{section_relaxed_pruning}

The pruning strategy described in Section~\ref{section_search_space_pruning} completely neglects combinations of labels with similar, but not equal, heuristic values. As illustrated by the example given in Section~\ref{section_motivation}, when pruning according to decomposability, single-label heads with marginally greater heuristic values are preferred to multi-label heads rated slightly worse. Relaxed pruning aims at tolerating minor declines in terms of a rule's heuristic value in favor of greater coverage. By relaxing the pruning constraints, and hence introducing a bias towards multi-label heads, more expressive rules are expected to be learned.

The main challenge of introducing such a bias revolves around two questions: First, the desired degree of the bias is unclear, i.e., how much of a decline in the heuristic value is tolerable. Second, the ideal number of labels in the head is unknown --- especially if rules may also predict the absence of labels. As both factors highly depend on the data set at hand, providing any recommendations is difficult. Moreover, the training time potentially suffers from relaxed pruning, as more label combinations are taken into account.

\subsection{Lifting the Heuristic Values of Rules}

We introduce a bias towards multi-label heads by multiplying the heuristic value $h$ of a rule with a dynamic weight $l \in \mathbb{R}$, which we refer to as a \emph{relaxation lift}. To prefer larger multi-label heads $l$ must increase with the number of labels in the head. The relaxation lift, which we will simply refer to as \emph{lift} in the remainder of this work, therefore specifies the decline in a rule's heuristic value that is acceptable in favor of predicting more labels.

To specify a relaxation lift for every number of labels $x \in \left[ 1, n \right]$ possibly contained in a head, we use \emph{relaxation lift functions} $\rho : \mathbb{R}_{+} \rightarrow \mathbb{R}$ mapping a given number of labels $x$ to a relaxation lift $l$. Although the function is only applied to natural numbers, defining $\rho$ in terms of real numbers facilitates the definition.

Given a rule $\boldsymbol{r}: H \leftarrow B$ and a lift function $\rho$, the \emph{lifted heuristic value} of the rule can be calculated as
\begin{equation}
\hat{h} = h \cdot \rho \left( x \right)
\end{equation}
where $x = \left| H \right|$ corresponds to the number of labels in the rule's head and $h$ is the (normal) heuristic value of the rule as calculated using a certain evaluation function (cf. Section~\ref{section_heuristics}). An example of how to calculate lifted heuristic values $\hat{h}$ is given in Table~\ref{table:lifted_heuristic_value}. These values are meant to be used as a replacement of the (normal) heuristic values $h$ when searching for multi-label heads (cf.  Section~\ref{section_search_space_pruning} and Figure~\ref{figure_example}).

\subsection{Relaxation Lift Functions}

\begin{figure}[t]
\begin{minipage}[b]{0.45\linewidth}
  \centering
  \begin{tabular}{c|c|c|c}
    $|H|$ & \textit{h} & $\rho(|H|)$ & $\hat{h}$\\
    \hline
    1 & 0.70 & $1.00$ & $0.70 \cdot 1.00 = 0.7000$\\
    2 & 0.67 & $1.07$ & $0.67 \cdot 1.07 = 0.7169$\\
    3 & 0.63 & $1.12$ & $0.63 \cdot 1.12 = 0.7056$\\
  \end{tabular}
  \captionof{table}{Example of calculating the lifted heuristic value by multiplying the normal heuristic value and the relaxation lift.}
  \label{table:lifted_heuristic_value}
\end{minipage}
\hspace{0cm}
\begin{minipage}[b]{0.55\linewidth}
  \centering
  \centerline{\includegraphics[scale=0.4]{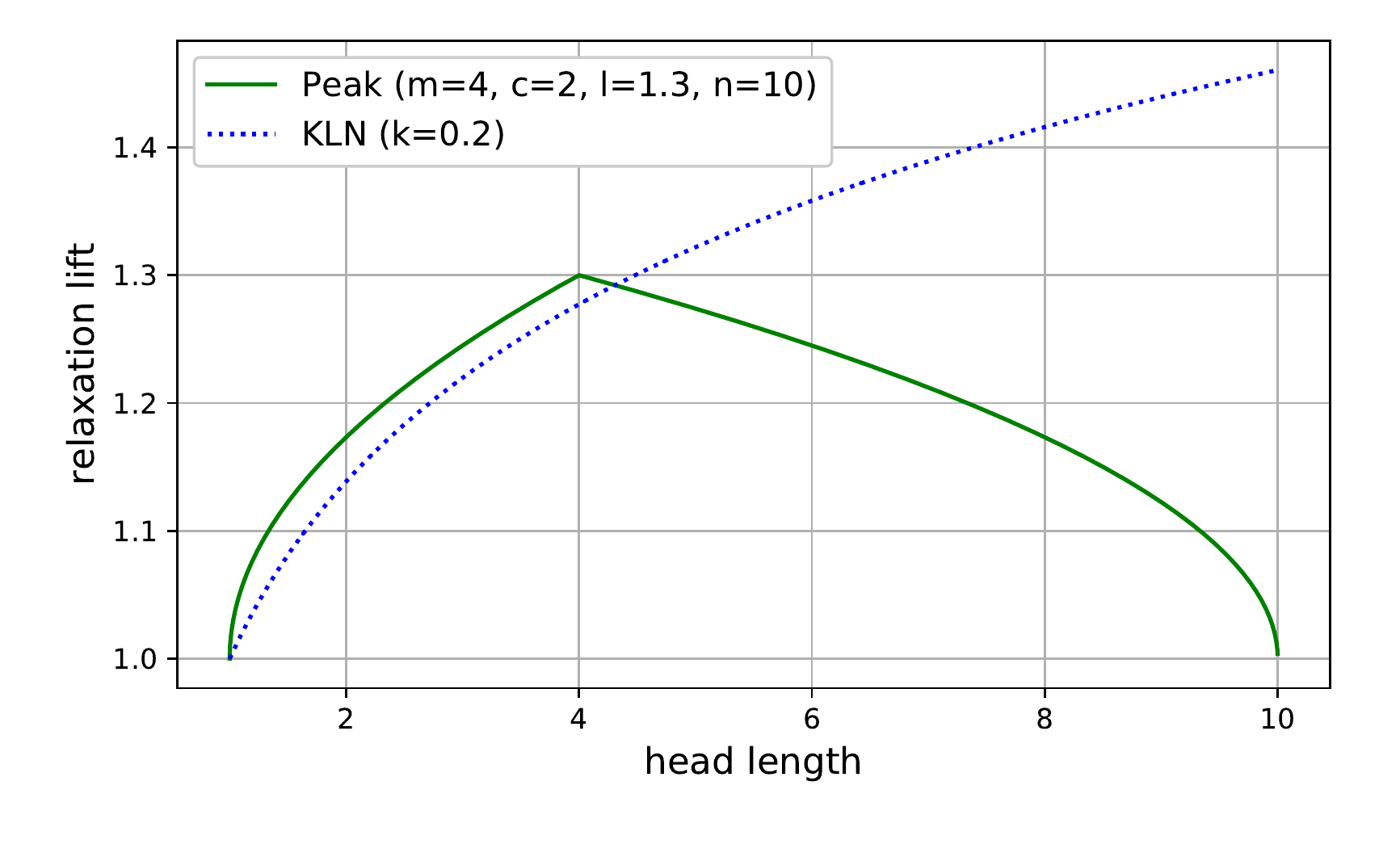}}
  \vspace{-0.5cm}
  \captionof{figure}{The KLN and peak relaxation lift functions.} \label{figure:kln_relaxation_lift_function}
\end{minipage}
\end{figure}

The proposed framework for relaxed pruning flexibly allows to utilize different relaxation lift functions with varying characteristics and effects on the rule induction process. In the following, we discuss two different types of functions used in this work. A visualization of these functions is given in Figure~\ref{figure:kln_relaxation_lift_function}.

\subsubsection{KLN Relaxation Lift Function.}

This simple lift function calculates as the natural logarithm of the number of labels $x$, multiplied by a user-customizable parameter $k \geq 0$. 
Adding an offset of $1$ to the calculated lift ensures that $l = 1$ in case of single-label heads. 
\begin{equation}
\rho_{\textit{KLN}} \left( x \right) = 1 + k \cdot \ln(x)
\end{equation}

The extent of the lift increases with greater values for the parameter $k$. Due to the natural logarithm, the function becomes less steep as the number of labels increases. This is necessary to prevent the selection of heads with a very large number of labels.

\subsubsection{Peak Relaxation Lift Function.}

This function also aims at preventing too many labels from being included in the head. With increasing number of labels $1, \dots, m$, where $m$ is a configurable parameter referred to as the \emph{peak}, the lift becomes greater, then decreases. This enables to introduce a bias towards heads with a specific number of labels, as they are lifted more than others. Given the peak $m$, the desired lift at the peak $l_{\textit{max}}$, the total number of available labels $n$, and a parameter $c$ that determines the curvature, the peak lift function is defined as follows. Note that $c = 1$ corresponds to a linear gradient.

\begin{gather}
\rho_{\textit{peak}} \left( x \right) = \begin{cases} 
  f_{m,1} \left( x \right) & \text{ if } x \leq m \\ 
  f_{m,n} \left( x \right) & \text{ otherwise}
\end{cases}\\
f_{a,b} \left( x \right) = 1 +  \left( \frac{x - b}{a - b} \right)^{\frac{1}{c}} \cdot \left(l_{\textit{max}} - 1\right)
\end{gather}

The advantage of the peak lift function is its efficiency, as more extensive pruning can be performed when using small values for the peak $m$.
Compared to the KLN lift function, it is less susceptible to including too many labels in the heads. 
Moreover, the peak lift function can be adapted more flexibly via the parameters $m$, $l_{\textit{max}}$, and $c$. However, as these parameters tend to have a significant impact on the learned model, this flexibility comes with a greater risk of misconfiguration.

\section{Relaxed Pruning of the Label Search Space}
\label{section:relaxed_search_space_pruning}

As we assess the quality of potential heads in terms of their lifted heuristic value $\hat{h}$, rather than $h$, it is necessary to adjust the search through the label space. In the following, we show that strictly pruning according to decomposability, as suggested in \cite{rapp2018}, does not yield the best head in terms of $\hat{h}$. Hence, we propose \emph{relaxed pruning} as an alternative and discuss the necessary changes in detail. We also provide an example that illustrates our approach.

\subsubsection{Suboptimal Pruning.}
When pruning according to decomposability, the best (multi-label) head is obtained by combining all single-label heads that reach the best heuristic value (cf. Figure \ref{figure_example}). By giving a simple counter-example, we show that this is not possible when searching for the head with the highest lifted heuristic value. Consider two heads \{$\hat{y}_1$\} and \{$\hat{y}_2$\} with (macro-averaged) heuristic values $0.8$ and $0.75$, respectively. As we do not lift single-label heads, the lifted and normal heuristic values are identical. Exclusively employing decomposability for finding the best performing lifted head results in the head \{$\hat{y}_1$\}, because combining both heads yields a lower value $0.775$. However, assuming the lift for two labels is $1.1$, the lifted heuristic value evaluates to $\hat{h} = 0.775 \cdot 1.1 = 0.8525$. Consequently, combining both heads results in a higher lifted heuristic value in such case. As a result, we conclude that the search space pruning suggested by Rapp~et~al.~\cite{rapp2018} is not suited to find the best head in terms of its lifted heuristic value $\hat{h}$. 

\begin{scriptsize}
\begin{algorithm}
\begin{algorithmic}[1]
\Procedure{FindBestHead}{$\emptyset \leftarrow B$}
\State S := $\{\hat{y} \leftarrow B : \hat{y} \in \hat{Y}\}$ \Comment{sorted single label heads}
\State $c := \emptyset \leftarrow B$; $c_\textit{best} := c$ \Comment{current candidate and best lifted candidate}
\For{$1,\dots,n$} \Comment{for all head lengths}
	\State $c = \Call{RefineCandidate}{S, c}$ \Comment{add best remaining label to head}
	\NoEndIf{$c.\hat{h} \geq c_\textit{best}.\hat{h}$} \Comment{update best lifted head}
		\State $c_\textit{best} = c$
	\NoEndIf{\Call{Prunable}{$c_\textit{best}, c$}} \Comment{check boundary}
		\State \textbf{return} $c_\textit{best}$ \Comment{return rule if TP $\geq$ FP}
\EndFor
\State \textbf{return} $c_\textit{best}$ \Comment{return rule if TP $\geq$ FP}
\EndProcedure
\end{algorithmic}
\caption{Search for the multi-label head with the greatest lifted heuristic value.}
\label{algo:decomposable}
\end{algorithm}
\end{scriptsize}

\subsubsection{Relaxed Pruning for macro-averaged Measures.}

We adjust the algorithm described in Section \ref{section_search_space_pruning} based on two observations: First, the best lifted head of length $k$ results from applying the lift to the head with the highest normal heuristic value of length $k$. As all heads of length $k$ are multiplied with the same lift, a head of length $k$ with a worse heuristic value cannot possibly achieve a higher lifted heuristic value. Thus, we obtain the best lifted head of a certain length by finding the best unlifted head.
Second, in case of decomposable evaluation measures that are computed via label-based macro-averaging, such as Hamming accuracy and macro-averaged F-measure (cf. Section~\ref{section_bipartition_functions} and \ref{section_heuristics}), we can guarantee that the best unlifted head of length $k$ results from combining the $k$ best single-label heads.

The basic structure of our algorithm is illustrated in Algorithm~\ref{algo:decomposable}. Similar to pruning according to decomposability, we need to evaluate all single-label heads on the training set for a given rule body and evaluation function (cf. solid line in Figure \ref{figure_example}). In accordance with our observations, we start with an empty head and successively add the best remaining single-label head (cf. \textsc{RefineCandidate} in Algorithm~\ref{algo:decomposable}). Using this strategy, we obtain the best unlifted multi-label head for each head length. We can then apply the lift in order to get the lifted heuristic value. During this process, we keep track of the head with the best lifted heuristic value $\hat{h}_\textit{best}$. When using a decomposable evaluation measure, including Hamming accuracy, we do not need to reevaluate any multi-label heads on the training data but can calculate their normal heuristic value as the average of the single-label heads' heuristic values.

Instead of generating each possible multi-label head, we calculate an upper bound $\hat{h}_\textit{upper}$ of the lifted heuristic value that could still be achieved by larger multi-label heads. For this, we multiply the normal heuristic value of the current head with length $k$ by the highest remaining lift, i.e., $\hat{h}_\textit{upper} = h_k \cdot \max_{k < i \leq n} \rho(i)$. If $\hat{h}_\textit{upper} < \hat{h}_\textit{best}$, we can prune as the highest performance cannot be achieved by longer heads (cf. \textsc{PRUNABLE} in Algorithm~\ref{algo:decomposable}). This results from the fact that the normal heuristic value cannot increase by adding more single-label heads as we start with the best. Thus, upper bounds $\max_{k < i \leq n} \rho(i)$ of the lift and the heuristic value $h_k$ are multiplied in order to obtain an upper bound $\hat{h}_\textit{upper}$ for the lifted heuristic value. This approach still guarantees finding the best performing lifted head for macro-averaged heuristics, i.e., also for the measures we are particularly interested in this work, namely macro F-measure and Hamming accuracy.

\subsubsection{(Approximate) Relaxed Pruning for decomposable Measures.}

Even though micro-averaged evaluation measures, such as the micro-averaged F-measure, are often decompoasable, combining the best $k$ single-label heads does not necessarily result in the best unlifted head of length $k$ in such case. This is, because the labels are not weighted equally as it is the case for macro-averaged measures. As a consequence, we cannot guarantee to find the best lifted head. Instead, we consider the introduced strategy for finding the best head of length $k$ as an approximation.
According to our experiments, this approximation seems to work well in practice --- most likely because we relax the search for optimal heuristic values anyway. 

\subsubsection{Complexity.} Compared to the original algorithm as described in Section~\ref{section_search_space_pruning}, the use of relaxed pruning does not require any additional evaluations of rules on the training instances. The number of evaluations on the training instances is proportional to the number of labels $n$ (multiplied by the number of training instances $m$) --- the same as for the original approach. However, in the worst case, it additionally requires to construct and evaluate $n - 1$ multi-label heads (cf. outer left path in Figure~\ref{figure_example}). However, as these heads can be evaluated based on the confusion matrices of the corresponding single-label heads, these additional steps are computationally cheap and do not require any additional evaluation on the training instances.

\subsubsection{Example.}
In this example, we illustrate the pruning procedure.
Suppose we use the KLN lift function with $k = 0.14$ for the example depicted in Figure \ref{figure_example}. Then $\rho(2) = 1.1$, $\rho(3) = 1.15$ and $\rho(4) = 1.19$. As mentioned, we follow the outer left path. For the head \{$\hat{y}_{1}$\} the lifted heuristic value and the maximum lifted value evaluate to $\hat{h} = \frac{2}{3} = \hat{h}_\textit{best}$ and $\hat{h}_{upper} = \frac{2}{3} \cdot 1.19 = 0.793$.\footnote{We round to three decimal places.} Because $\hat{h}_\textit{upper} \geq \hat{h}_\textit{best}$, we cannot prune at this point. For the head \{$\hat{y}_{1}, \hat{y}_{2}$\}, $\hat{h}_\textit{upper}$ stays the same, but the lifted heuristic value evaluates to $\hat{h} = \frac{2}{3} \cdot 1.1 = 0.733 = \hat{h}_\textit{best}$. As $\hat{h}_\textit{upper} \geq \hat{h}_\textit{best}$, we still need to check the head \{$\hat{y}_{1}, \hat{y}_{2}, \hat{y}_{3}$\}, for which we calculate $\hat{h} = \frac{5}{9} \cdot 1.15 = 0.639$ and $\hat{h}_{upper} = \frac{5}{9} \cdot 1.19 = 0.661$. As the pruning criterion $\hat{h}_\textit{upper} < \hat{h}_\textit{best}$ holds, we terminate the search and return the best head. The dashed line in Figure \ref{figure_example} indicates which heads need to be examined when using relaxed pruning. Note that the best lifted and unlifted head are the same in this example.

\subsubsection{Fixing the Head.}
During the rule refinement process rules are specialized by adding additional conditions to the body. When searching for a new (multi-label) head each time a rule has been modified, as suggested in \cite{rapp2018}, previously found heads are often discarded in favor of single-label heads with lower coverage but a higher (lifted) heuristic value. However, keeping the original head and modifying the body accordingly might result in a better rule. To address this problem, we fix the head during the rule refinement process, i.e., we keep the original head instead of searching for a new head each time the body is modified. As a positive side effect of this modification the time required for building a model usually decreases as it is not necessary to frequently search for new heads.

\subsubsection{Constraints on Rules.}
In addition to fixing the head, as discussed in the previous section, we require each rule to predict at least as many $\tp$ as $\fp$, which effectively imposes a lower bound on the quality of the rules. In preliminary experiments, we found this constraint to be helpful to prevent suboptimal label predictions from being included in the heads for the sake of increasing its lift.

Moreover, we require each label assignment in a head to result in at least one $\tp$. This prevents label assignments that do not affect the (normal) heuristic value of a rule, but would result in a higher lift, from being added to a head. For example, such situation might occur if a label is already predicted for all training instances by previously induced rules.

\begin{figure}[b!]
\centering
\begin{minipage}{0.66\textwidth}
  \centering
  \includegraphics[width=0.85\textwidth]{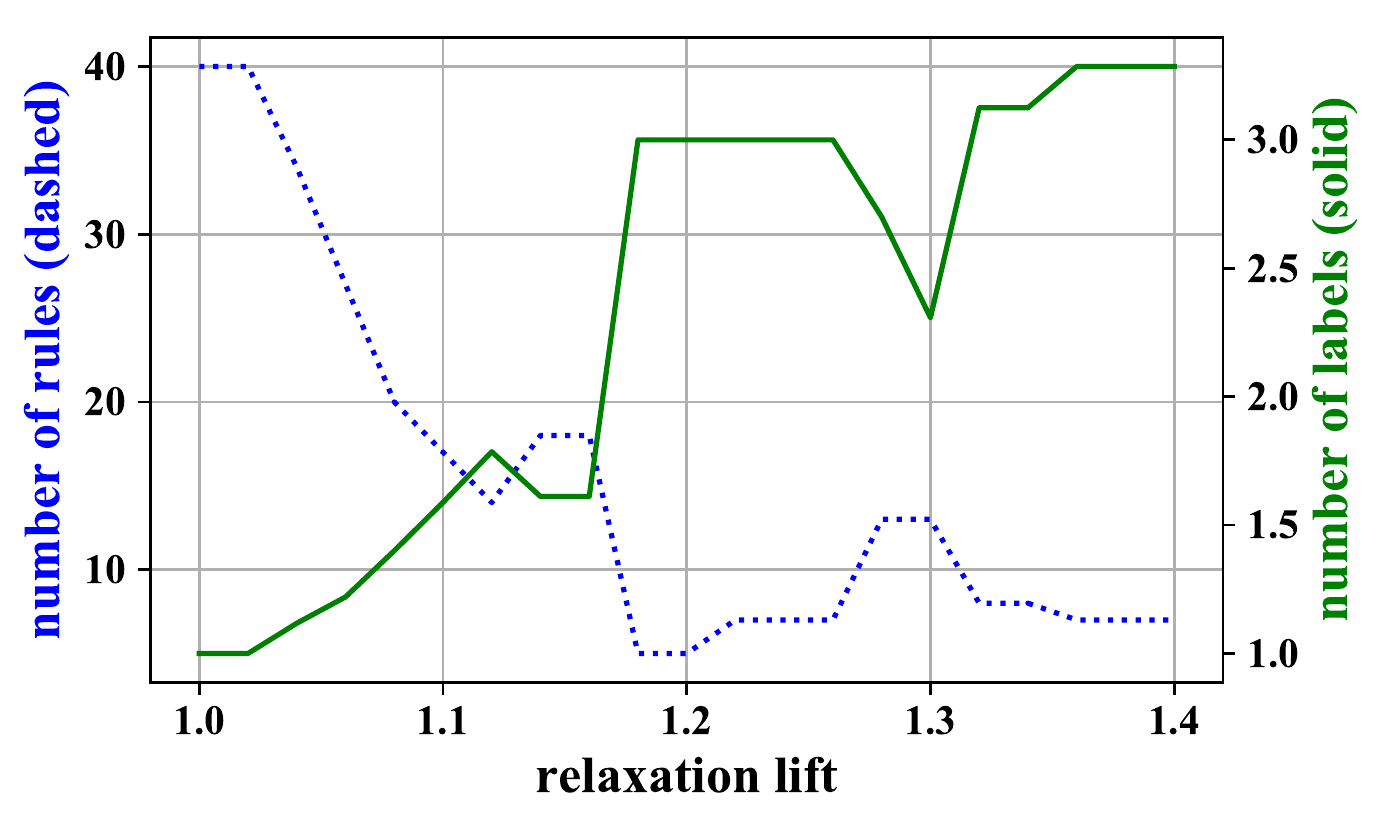}
  \vspace{-0.3em}
  \captionof{figure}{Sensitivity analysis for the KLN lift function on \textsc{flags}.}
  \label{figure:sensitivity_analysis}
\end{minipage}
\begin{minipage}{0.33\textwidth}
  \centering
  \includegraphics[width=1.0\textwidth]{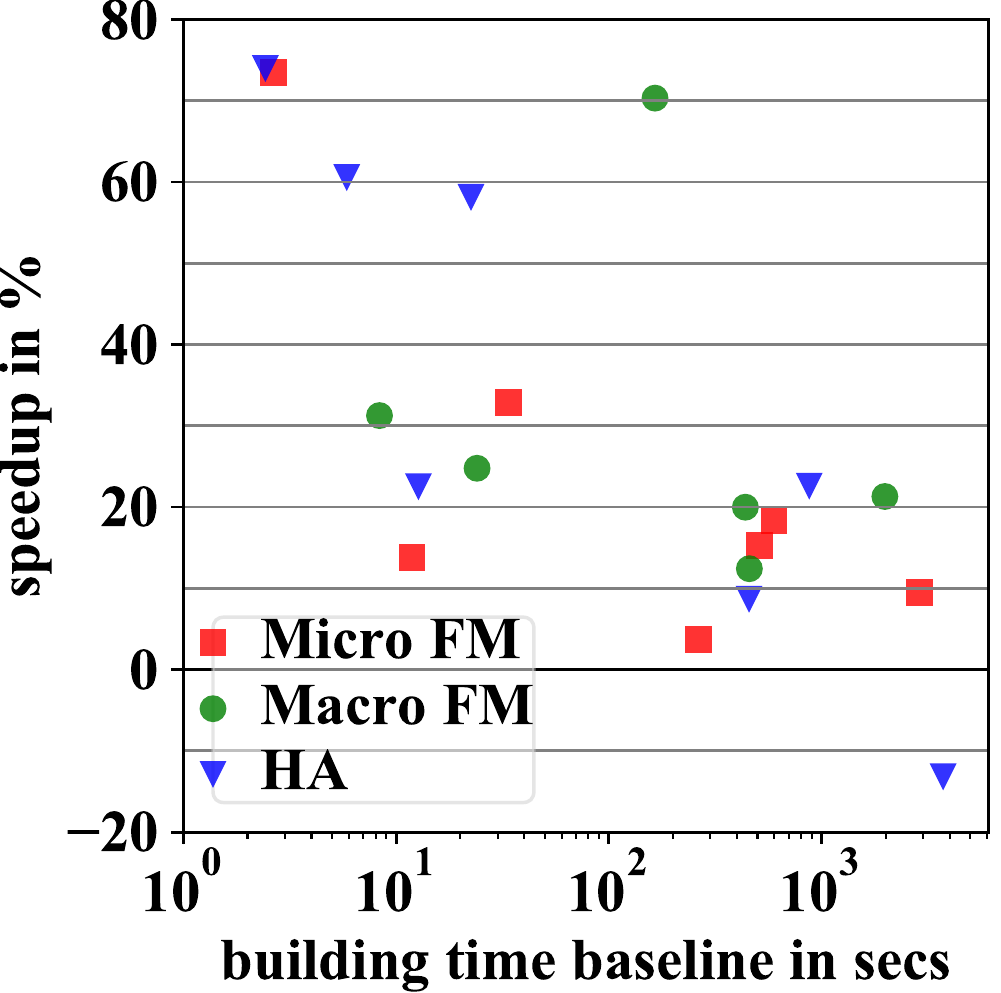}
  \captionof{figure}{Time comparison.}
  \label{figure:computational_costs}
\end{minipage}
\end{figure}

\section{Evaluation}
\label{section_evaluation}

In this section, we demonstrate the effectiveness of our approach empirically. For our analysis, we consider the predictive performance, the model characteristics, as well as the training time, and give examples of multi-label rules learned by our method.

\subsubsection{Experimental Setup.}
We tested our method using relaxed pruning on seven multi-label data sets and compared it to the approach by Rapp~et~al.~\cite{rapp2018} using the same configuration.\footnote{We used the data sets \textsc{birds}, \textsc{flags}, \textsc{cal500}, \textsc{emotions}, \textsc{medical}, \textsc{scene} and \textsc{yeast} from \url{http://mulan.sf.net/datasets-mlc.html}. The source code and data sets are publicly available at \url{https://github.com/keelm/SeCo-MLC/tree/relaxed-pruning}} To isolate the influence relaxed pruning has on the learned models, we transferred our additions to fix the heads and impose constraints on the learned rules as discussed earlier. For every data set and every target performance measure (HA, macro- and micro-averaged FM with $\beta=0.5$, as suggested by \cite{rapp2018}), we determined the best lift setting from a number of candidates using five-fold cross-validation on the training set. If two lift settings achieved the same performance, we chose the one with a higher lift as it will typically result in a more compact model. After evaluating all candidates, we trained a model using the best setting and validated it on the test set. We denote the ability to learn rules that predict the presence and absence of labels as $+$ and $-$, respectively. Further, we abbreviate micro-averaging (mic or micro) and macro-averaging (mac or macro). \emph{Subset accuracy} (SA) measures the percentage of perfectly classified examples.

\subsubsection{Sensitivity Analysis and Model Characteristics.}
Figure~\ref{figure:sensitivity_analysis} depicts the number of rules and the average number of labels in their heads, depending on the extent of the lift that results from using the KLN lift function and macro-averaged FM.
As expected, greater lifts tend to result in larger heads and fewer rules.
This is, because rules that predict several labels, rather than a single one, have greater coverage.
As a consequence, fewer rules are required to cover the entire training data.
However, if the lift is too high, very generic rules, which predict the majority class label and are unable to model the training data accurately, are learned.

Moreover, we observe that the average number of conditions in the bodies decreases with higher lifts. As we induce rules with larger multi-label heads, we would have expected label conditions to be used more frequently.
Surprisingly, the percentage of label conditions approaches zero even for a moderate lift. 
For the peak lift function, the overall trend is identical. However, the maximum number of labels in the head is typically limited.

In addition to the sensitivity analysis, we list some characteristics of models learned during the evaluation in Table~\ref{table:model_characteristics}. We can observe similar phenomena as in the sensitivity analysis. The model characteristics, however, show that our observations also seem to hold for the best lift setting.

\subsubsection{Computational Costs.}

As shown in Figure~\ref{figure:computational_costs}, we compare the training times of our method to the baseline by Rapp~et~al.~\cite{rapp2018} using the same configuration. The horizontal axis corresponds to the times required by the baseline to build the models. The vertical axis denotes the relative speedup (or slowdown) that results from using relaxed pruning.
Although it potentially evaluates more heads, our method is faster in most of the cases. Typically, a speedup between 10\% and 25\% can be observed.
As we isolate relaxed pruning from the other changes, the speedup most likely results from fewer rules being learned due to their increased coverage as discussed above. 
Furthermore, we observe that the average number of conditions per rule decreases when using relaxed pruning, i.e., fewer refinement candidates must be taken into account.

\newcommand{\win}[1]{\textbf{#1}}
\begin{figure}[h]
\begin{minipage}{0.60\textwidth}
\resizebox{1.0\textwidth}{!}{
\begin{minipage}{1.02\textwidth}
\begin{scriptsize}
\begin{tabular}{C{1.75cm} | c  c | c  c | c  c | c  c | c  c | c c }
                     Number of    & \multicolumn{2}{c|}{Mic FM$^+$} & \multicolumn{2}{c|}{Mic F$^+_-$} & \multicolumn{2}{c|}{Mac F$^+$} & \multicolumn{2}{c|}{Mac F$^+_-$} & \multicolumn{2}{c|}{HA$^+$} & \multicolumn{2}{c}{HA$^+_-$}\\ \hline
Rules                  & 140 & 140 & 132 & 92 &140 & 129 & 132 & 113 & 162 & 136 & 58 & 23\\ \hline
Conditions             & 219 & 213 & 184 & 146 & 220 & 199 & 184 & 175 & 254 & 204 & 58 & 29\\ \hline
Label conditions        & 7 & 4 & 1 & 2 & 7 & 3 & 1 & 1 & 3 &0 & 1 & 0\\ \hline
Multi-label heads         & 1 & 5 & 0 & 22 & 1 & 8 & 0 & 14 & 1 & 30 & 0 & 12\\ \hline
Labels per multi-label head  & 2.0 & 2.0 & - & 2.59 & 2.0 & 2.0 & - & 2.57 & 2.0 & 2.1 & - & 17.0\\ 
\end{tabular}
\end{scriptsize}
\end{minipage}
}
\captionof{table}{Model characteristics for \textsc{birds}. For each heuristic the left and right column shows the values for the normal and relaxed pruning approach, respectively.}
\label{table:model_characteristics}
\end{minipage}
\hspace{0.01\textwidth}
\begin{minipage}{0.35\textwidth}
  \centering
  \begin{scriptsize}
\begin{tabular}{l | C{0.52cm} | C{0.52cm} | C{0.52cm} | C{0.52cm}}
Heuristic& HA       & Mic FM    &Mac FM & SA \\ \hline
HA$^+$        & 2/\win{4}/1    & 1/\win{5}/1         & 2/\win{4}/1 & 0/\win{4}/3\\
HA$^+_-$      & \win{5}/1/1        & \win{4}/2/1         & 2/\win{4}/1    & \win{5}/0/2\\ \hline
Mic FM$^+$    & 3/3/1        & 3/3/1         & 3/3/1    & \win{3}/2/2\\
Mic FM$^+_-$  & \win{3}/1/3      & 1/\win{3}/3         & 1/\win{3}/3    & 2/2/3  \\ \hline
Mac FM$^+$    & 1/\win{5}/1        & 1/\win{5}/1         & 3/3/1    & 2/\win{3}/2\\
Mac FM$^+_-$  & \win{2}/1/4       & \win{2}/1/4         & \win{2}/1/4    & \win{2}/1/4 \\ 
\end{tabular}
\end{scriptsize}
\vspace{-2pt}
\captionof{table}{Number of wins / losses / ties of relaxed pruning.}
  \label{table:predictive_performance}
\end{minipage}
\end{figure}

\subsubsection{Predictive Performance.}
Table~\ref{table:predictive_performance} lists the number of wins and losses of the compared approaches.
We conclude that relaxed pruning results in a predictive performance that is comparable to that of the baseline, despite learning more compact models. More precisely, we observe a decline in performance when using HA$^+$ or macro FM$^+$ as the objective for inducing the heads, but using macro FM$^+_-$ or, in particular, HA$^+_-$ results in an improvement.
For micro FM$^+$, the performance is quite similar, despite missing the guarantees discussed in Section~\ref{section:relaxed_search_space_pruning}.
Regarding an overall comparison between all approaches and heuristics, we can observe that learning the absence and presence of labels and seeking for relaxed Hamming accuracy (HA R$^+_-$) ranks highest in average among the 12 approaches w.r.t. Hamming accuracy, 
but also for subset accuracy, which no approach dedicatedly addresses.
In contrast, for micro and macro F-measure, the best performing models are obtained by using relaxed pruning together with the micro F-measure and only predicting the presence of labels (Mic FM R$^+$). This reflects the focus of the F-measure on positive labels compared to HA.  
In conclusion, relaxing the pruning constraints and deliberately preferring rules with a worse heuristic value in favor of coverage and expressiveness does not seem to have a negative effect on the predictive performance of the models and even results in improvements in some cases.

As mentioned earlier, we determined the best lift settings on the training data. In the majority of the cases, the peak lift function is preferred to the KLN lift function. Due to the variety of possible parameter settings, the peak lift function is more difficult to tune. We observe a trend towards lifts clearly greater than 1. We assume that the parameter for specifying the lift mainly depends on the average number of labels per instance. Moreover, it may also be relevant whether the absence of labels is predicted by the rules in addition to the relevance of labels, as we expect that a greater peak might be beneficial in such case.

\begin{figure}[b!]
\begin{minipage}{1\textwidth}
\resizebox{1.0\textwidth}{!}{
\begin{minipage}{1.\textwidth}
\noindent\rule{\textwidth}{0.6pt}
\begin{tabular}{ L{4.6cm} l | L{4.52cm} l}
\labelfont{Class5} $\leftarrow$ Att61 & (112, 50) & \labelfont{Class4}, \labelfont{Class5} $\leftarrow$ Att61 & (230, 94)\\
\labelfont{Class4} $\leftarrow$ \labelfont{Class5} &(118, 44) & &\\ \hline
\labelfont{Class3} $\leftarrow$ Att50 & (84, 50) & \labelfont{Class2}, \labelfont{Class3} $\leftarrow$ Att50 & (174, 111)\\
\labelfont{Class2}  $\leftarrow$ \labelfont{Class3} & (146, 141) & &\\ 
\end{tabular}
\footnotesize
\noindent\rule{\textwidth}{0.6pt}
\begin{tabular}{ l l | l l}
\labelfont{red} $\leftarrow$ colours$_1$, area$_1$, bars, crescent & (85, 9) & \labelfont{red}, \labelfont{white} $\leftarrow$ colours$_1$ & (166, 38)\\
\labelfont{red} $\leftarrow$ bars, crescent, colours$_2$, area$_1$, stripes &(13, 6) & \labelfontinv{green}, \labelfont{red} $\leftarrow$ colours$_2$ & (71, 35)\\
\labelfontinv{red} $\leftarrow$ area$_2$, circles &(9, 1) & \labelfont{green}, \labelfontinv{red} & (2, 0)\\
\labelfont{red} $\leftarrow$ sunstars$_1$ &(4, 0)&&\\
\labelfontinv{red} $\leftarrow$ sunstars$_2$ &(1, 0)&&\\
\labelfont{red} &(1, 0)&&\\ 
\end{tabular}

\noindent\rule{\textwidth}{0.6pt}
\begin{tabular}{ l l | l l}
\labelfontinv{RBN} $\leftarrow$ ssd59 & (288, 0) & \labelfontinv{BHG}, \labelfontinv{Warbling Vireo}, \labelfontinv{MGW}, \labelfontinv{Stellar's Jay}, \labelfontinv{RBN} $\leftarrow$ ssd63 & (1012, 3)\\
\labelfontinv{RBN} $\leftarrow$ ssd89 & (21, 0) &\labelfontinv{MGW}, \labelfontinv{RBN} $\leftarrow$ ssd56 & (141, 0) \\
\labelfont{RBN} $\leftarrow$ ssd153 & (4, 0) & \labelfontinv{Common Nighthawk}, \labelfontinv{RBN} $\leftarrow$ ssd7, ssd145  & (190, 0)\\
\labelfontinv{RBN} & (9, 0) & \labelfontinv{RBN} $\leftarrow$ ssd8 & (16, 0)\\
& & \labelfont{RBN} $\leftarrow$ ssd45 & (4, 0) \\
& & \labelfontinv{RBN} & (1, 0)
\end{tabular}

\noindent\rule{\textwidth}{0.6pt}
\end{minipage}
}
\captionof{figure}{Selected learned normal (left) and relaxed (right) pruning rules regarding a specific label. We show ($\tp$, $\fp$) and absence of label \labelfont{x} as \labelfontinv{x}. Top down: \textsc{yeast} (macro FM$^+$) twice, \textsc{flags} (micro FM$^+_-$) and \textsc{birds} (macro FM$^+_-$). We abstract specific conditions and only represent attribute names, but indicate different values.
For \textsc{birds} we abbreviate Red-breasted Nuthatch (RBN), Black-headed Grosbeak (BHG), MacGillivray's Warbler (MGW) and audio-ssd (ssd).
}
\label{figure:rule_sets}
\end{minipage}
\end{figure}

\subsubsection{Exemplary Rules.}
In Figure~\ref{figure:rule_sets} we show exemplary rules as induced with and without the use of relaxed pruning. It can be seen that multi-label heads and label conditions are both suited to model label dependencies. Depending on the model, these different representations may even be equivalent in meaning (cf. first row). Whereas the use of relaxed pruning seems to result in fewer label conditions being learned, it often results in significantly more multi-label heads. This makes a quantitative analysis of the number of label dependencies discovered by the respective approaches more difficult. Nevertheless, our results suggest that relaxed pruning helps to model label dependencies in the form of multi-label heads. Such heads often provide a more compact representation of the discovered correlations. In contrast to label conditions, rules with multi-label heads provide useful information on their own. They do not require to take the order of the rules into account and must not be interpreted in the context of other rules. Due to these advantages, we argue that multi-label heads are easier to understand in many cases.

\section{Related Work}
\label{section_related_work}

Most approaches to multi-label rule learning found in the literature are based on association rule (AR) discovery. Alternatively, a few approaches use evolutionary algorithms or classifier systems for evolving multi-label classification rules \cite{RuleBasedMLCwLCM,EvoMLCuARM,MLCRules}. Inducing rules with several labels in the head is usually implemented as a post-processing step. For example, \cite{thabtah06MLassociative} and similarly \cite{MLCwARs} induce single-label association rules that are merged to create multi-label rules. By using a separate-and-conquer strategy the step of inducing descriptive but often redundant models of the data is omitted. Instead, classification rules that aimed at providing accurate predictions are learned directly \cite{menc15}.

Most of the approaches mentioned above are restricted to expressing a certain type of relationship since labels are only allowed in the heads of the rules. Approaches that also use labels as antecedents are often restricted to global label dependencies, such as the approaches by \cite{jf:PL-08-WS-Park,LI-MLC,papagiannopoulou15deterministicrelations} that use the relationships discovered by AR mining on the label matrix for refining the predictions of multi-label classifiers.

\section{Conclusions}
\label{section_conclusions}

In this work, we demonstrated the effectiveness of introducing a bias towards rules with larger multi-label heads. By deliberately preferring rules with a worse heuristic value, we are capable of learning more compact models with more expressive rules that are explicitly tailored to exploit label dependencies. 
In addition, we argued that strict upper bounds in terms of computational complexity still hold when using relaxed pruning and our experiments revealed that training time even tends to decrease due to the increased coverage of the induced rules. In general, we are able to achieve comparable predictive performance --- observing gains in performance for 3 out of 6 tested objectives.\\

\bibliographystyle{plain}
\bibliography{bibliography}

\begin{thebibliography}{10}

\bibitem{RuleBasedMLCwLCM}
Miltiadis Allamanis, Fani~A Tzima, and Pericles~A Mitkas.
\newblock Effective rule-based multi-label classification with learning
  classifier systems.
\newblock In {\em International Conference on Adaptive and Natural Computing
  Algorithms}, 2013.

\bibitem{EvoMLCuARM}
J~Arunadevi and V~Rajamani.
\newblock An evolutionary multi label classification using associative rule
  mining for spatial preferences.
\newblock {\em IJCA Special Issue on Artificial Intelligence Techniques-Novel
  Approaches and Practical Applications}, 2011.

\bibitem{MLCRules}
Jos{\'e}~Luis {\'A}vila-Jim{\'e}nez, Eva Gibaja, and Sebasti{\'a}n Ventura.
\newblock Evolving multi-label classification rules with gene expression
  programming: A preliminary study.
\newblock In {\em International Conference on Hybrid Artificial Intelligence
  Systems}, 2010.

\bibitem{LI-MLC}
Francisco Charte, Antonio~J Rivera, Mar{\'{\i}}a~Jos{\'{e}} del Jes{\'{u}}s,
  and Francisco Herrera.
\newblock {LI-MLC:} {A} label inference methodology for addressing high
  dimensionality in the label space for multilabel classification.
\newblock {\em {IEEE} Transactions on Neural Networks and Learning Systems},
  25(10), 2014.

\bibitem{demb12}
Krzysztof Dembczy{\'n}ski, Willem Waegeman, Weiwei Cheng, and Eyke
  H{\"u}llermeier.
\newblock On label dependence and loss minimization in multi-label
  classification.
\newblock {\em Machine Learning}, 88(1-2), 2012.

\bibitem{lakkaraju2016}
Himabindu Lakkaraju, Stephen~H Bach, and Jure Leskovec.
\newblock Interpretable decision sets: A joint framework for description and
  prediction.
\newblock In {\em International Conference on Knowledge Discovery and Data
  Mining}, 2016.

\bibitem{MLCwARs}
Bo~Li, Hong Li, Min Wu, and Ping Li.
\newblock {Multi-label Classification based on Association Rules with
  Application to Scene Classification}.
\newblock In {\em International Conference for Young Computer Scientists},
  2008.

\bibitem{MLRLbook}
Eneldo Loza~Menc{\'{\i}}a, Johannes F{\"{u}}rnkranz, Eyke H{\"{u}}llermeier,
  and Michael Rapp.
\newblock Learning interpretable rules for multi-label classification.
\newblock In {\em Explainable and Interpretable Models in Computer Vision and
  Machine Learning}. Springer, 2018.

\bibitem{menc15}
Eneldo~Loza Menc{\'\i}a and Frederik Janssen.
\newblock Learning rules for multi-label classification: A stacking and a
  separate-and-conquer approach.
\newblock {\em Machine Learning}, 105(1), 2016.

\bibitem{papagiannopoulou15deterministicrelations}
Christina Papagiannopoulou, Grigorios Tsoumakas, and Ioannis Tsamardinos.
\newblock Discovering and exploiting deterministic label relationships in
  multi-label learning.
\newblock In {\em ACM SIGKDD International Conference on Knowledge Discovery
  and Data Mining}, 2015.

\bibitem{jf:PL-08-WS-Park}
Sang-Hyeun Park and Johannes F{\"{u}}rnkranz.
\newblock Multi-label classification with label constraints.
\newblock In {\em ECML PKDD 2008 Workshop on Preference Learning}, 2008.

\bibitem{rapp2018}
Michael Rapp, Eneldo~Loza Menc{\'\i}a, and Johannes F{\"u}rnkranz.
\newblock Exploiting anti-monotonicity of multi-label evaluation measures for
  inducing multi-label rules.
\newblock In {\em Pacific-Asia Conference on Knowledge Discovery and Data
  Mining}, 2018.

\bibitem{thabtah06MLassociative}
Fadi~A Thabtah, Peter~I Cowling, and Yonghong Peng.
\newblock Multiple labels associative classification.
\newblock {\em Knowledge and Information Systems}, 9(1), 2006.

\bibitem{tsoumakas10MLoverview}
Grigorios Tsoumakas, Ioannis Katakis, and Ioannis Vlahavas.
\newblock Mining multi-label data.
\newblock In {\em Data Mining and Knowledge Discovery Handbook}. Springer,
  2009.

\end{thebibliography}

\end{document}